\documentclass{article}

\PassOptionsToPackage{numbers, sort&compress}{natbib}
\usepackage[preprint]{neurips_2026}

\usepackage[utf8]{inputenc}
\usepackage[T1]{fontenc}
\usepackage{hyperref}
\usepackage{url}
\usepackage{booktabs}
\usepackage{amsfonts}
\usepackage{nicefrac}
\usepackage{microtype}
\usepackage{xcolor}
\usepackage{multirow}
\usepackage{graphicx}
\usepackage{amsmath}
\usepackage{arydshln}
\usepackage{threeparttable}

\usepackage{cleveref}
\crefname{section}{Sec.}{Secs.}
\Crefname{section}{Section}{Sections}
\Crefname{table}{Table}{Tables}
\crefname{table}{Tab.}{Tabs.}
\Crefname{figure}{Figure}{Figures}
\crefname{figure}{Fig.}{Figs.}

\title{PIXLRelight: Controllable Relighting via Intrinsic Conditioning}

\author{%
  Miguel Farinha \quad Ronald Clark \\
  Department of Computer Science\\
  University of Oxford \\
  \texttt{\{miguel.farinha,ronald.clark\}@cs.ox.ac.uk}
}

\begin{document}

\maketitle

\begin{abstract}
We present \textsc{PIXLRelight}, a feed-forward approach for physically controllable single-image relighting. Existing methods either provide limited lighting control (e.g.\ through text or environment maps), accumulate errors when chaining inverse and forward rendering, or require costly per-image optimization. Our key idea is to bridge physically based rendering (PBR) and learned image synthesis through a shared intrinsic conditioning that can be obtained from either real photographs or PBR renders. At training time, paired multi-illumination photographs are decomposed into albedo, diffuse shading, and non-diffuse residuals, which condition the model. At inference time, the same conditioning is computed from a path-traced render of a coarse 3D reconstruction of the input under user-specified PBR lights. A transformer-based neural renderer then applies the target illumination to the source photograph, preserving fine image detail through a per-pixel affine modulation. \textsc{PIXLRelight} enables arbitrary PBR-style lighting control, achieves state-of-the-art relighting quality, and runs in under a tenth of a second per image. Code and models are available at https://mlfarinha.github.io/pixl-relight/.
\end{abstract}

\begin{figure}[!ht]
  \centering
  \includegraphics[width=\linewidth]{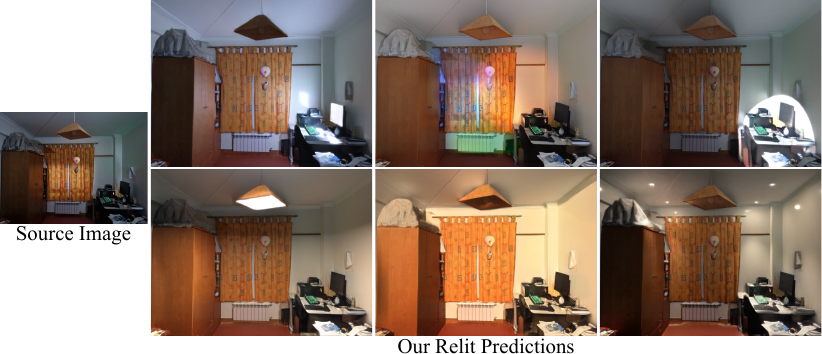}
  \caption{\textbf{\textsc{PIXLRelight}} is a feed-forward transformer that produces photorealistic relightings of in-the-wild photographs in a single pass. The user authors the target illumination in a physically based renderer, with full control over light type, position, color, and intensity. The model is conditioned on the resulting intrinsic decomposition of the target appearance -- albedo, diffuse shading, and a non-diffuse residual. Above, the same source image is relit under six different illuminations.}
  \label{fig:banner}
\end{figure}

\section{Introduction}
\label{sec:introduction}

Illumination is a primary component of image formation. In computer graphics, physically based rendering (PBR) engines such as Blender~\citep{blender} and Unreal Engine~\citep{unreal} expose lighting controls directly: an artist places point, area, directional, environmental, or emissive light sources in a 3D scene, and a path tracer simulates their interaction with geometry and materials. Bringing this same physical control to in-the-wild photographs would unlock applications across computational photography, content creation, and visual effects -- but is fundamentally harder, since recovering the geometry, materials, and light transport from an image is challenging.

Recent approaches address this gap in three ways, none of which jointly enables physical control, photorealism, and speed. A first line conditions relighting on HDR environment maps~\citep{jin2024neuralgaffer, zeng2024dilightnet}, reference images~\citep{xing2025luminet}, or screen-space scribbles~\citep{choi2025scribblelight} and masks~\citep{magar2025lightlab}; these interfaces struggle to express spatially localized, multi-source illumination. A second line treats relighting as an inverse-then-forward rendering pipeline that estimates G-buffers and re-renders them under a target illumination~\citep{zeng2024rgbx, liang2025diffusionrenderer, sun2025ouroboros, fang2025vrgbx}; the intermediate buffers cannot encode every cue the renderer needs (e.g.\ transparency, subsurface scattering), and errors compound across the two stages~\citep{he2025unirelight}. The third, and closest, line combines PBR with a neural renderer:~\citet{careaga2025physically} reconstruct an approximate textured mesh from a photograph, ray-trace it under user-specified illumination, and pass the CG render to a neural renderer that produces the final image. Two limitations follow. First, because the mesh is reconstructed with diffuse reflectance only, the ray-traced render is diffuse-only, and the network must guess every specular highlight, transparent surface, or refractive cue in the output. Second, training each input pair requires a per-image differentiable-rendering optimization that fits a 3D lighting environment to the source image; this optimization is stable only for a narrow lighting parameterization, bounding the lighting distribution the network sees at training.

The natural alternative is to decompose the photograph into geometry and materials, place new lights in PBR, and re-render. This is exactly what computer graphics does well -- given high-quality assets. For in-the-wild photographs, those assets are precisely what we cannot recover reliably: single-image geometry is coarse and material decomposition is under-constrained, so a path-traced render of such a reconstruction is riddled with artifacts. Our key insight is to harness both the advantages of PBR (controllability) and neural rendering (detailed photorealism). Our insight is to use each tool for what it does well: PBR for specifying \emph{what} the target lighting should be, and a feed-forward neural model trained on real photographs for \emph{how} to apply that lighting to the photograph photorealistically -- absorbing the imperfections of the underlying scene reconstruction in the process. The two are bridged by an intrinsic decomposition of the target appearance into albedo, diffuse shading, and a non-diffuse residual~\citep{ke2025marigold}. The same input is produced from a real photograph at training and from a PBR render at inference. At training, paired captures from existing multi-illumination datasets~\citep{miiw, bigtime, vidit} pass through a frozen intrinsic decomposition model~\citep{ke2025marigold} to directly supervise relighting which means there is no inverse-then-forward chain to train or per-image rendering optimization to run. At inference, the conditioning is produced by a path-traced render of a coarse 3D reconstruction in which the user freely controls lighting using arbitrary combinations of physically-based lights. Crucially, the model never sees the rendered RGB image; it sees only the intrinsic buffers, which carry a precise cue for the desired output lighting.

We call this approach \textsc{PIXLRelight}, a feed-forward transformer that consumes a source image and the target intrinsics and predicts a relit RGB image with per-pixel detail. Following recent feed-forward dense-prediction architectures~\citep{wang2025vggt, jiang2025rayzer, lin2025depthanything3, jin2024lvsm}, we tokenize the two inputs with asymmetric encoders -- a ViT~\citep{dosovitskiy2020vit} for the source image and a ConvNeXt~\citep{liu2022convnext} for the smoother, lower-frequency intrinsic stack -- and fuse them in a shared transformer trunk read out by a DPT head~\citep{ranftl2021dpt}. Rather than regressing RGB directly, the head produces an identity-initialized per-pixel affine modulation of the source: at initialization the network reproduces the input exactly, and during training it learns only the residual lighting transformation, preserving photorealistic detail by construction.

In summary, our contributions are: (1) a target-appearance intrinsic decomposition as a single conditioning interface for single-image relighting, computed from a real image at training and from a PBR render at inference; (2) a direct training strategy supervised end-to-end on real multi-illumination photographs, without an inverse-then-forward rendering pipeline or per-image rendering optimization; and (3) \textsc{PIXLRelight}, which achieves state-of-the-art relighting quality while running in under a tenth of a second per image.
\section{Related work}
\label{sec:related}

\paragraph{Single-image relighting} methods can be organized by the modality through which the user specifies the target lighting. Text and reference-image conditioning~\citep{zhang2025iclight, xing2025luminet} hallucinate plausible illumination but offer no physical control. Environment-map conditioning~\citep{jin2024neuralgaffer, zeng2024dilightnet, liang2025diffusionrenderer, he2025unirelight} assumes lighting comes from infinity and cannot express the near-field sources common in indoor scenes. Parametric conditioning on a single light source placed in a reference view, as in GenLit~\citep{bharadwaj2025genlit} and SyncLight~\citep{serrano2026synclight}, restores spatial control but is not designed for multi-light, area-light, or emissive-geometry edits. Pixel-aligned intrinsic conditioning sidesteps both limitations: RGB$\leftrightarrow$X~\citep{zeng2024rgbx}, Ouroboros~\citep{sun2025ouroboros}, and V-RGBX~\citep{fang2025vrgbx} drive diffusion-based renderers from intrinsic buffers that include shading, but stop short of CG-style authoring. Closest to ours,~\citet{careaga2025physically} achieve Blender-authored physical control by routing a PBR render through a feed-forward neural renderer, but their pipeline requires a per-image differentiable-rendering optimization for training and a diffuse-only RGB render as conditioning. \textsc{PIXLRelight} shares the Blender-as-authoring-interface design but conditions the network on rendered intrinsic fields rather than rendered RGB, and trains directly on paired multi-illumination captures.

\paragraph{Intrinsic image decomposition} factors an RGB image into illumination-invariant material properties and illumination-dependent shading~\citep{barrow1978intrinsics}. Recent diffusion-based decompositions~\citep{kocsis2024iid, zeng2024rgbx, fang2025vrgbx, ke2025marigold, kocsis2025intrinsix} substantially improve over earlier hand-crafted~\citep{land1971lightness, grosse2009ground, careaga2023ordinal} and supervised~\citep{li2020inverse, wang2021inverse, zhu2022irisformer, li2021openrooms} approaches. We adopt Marigold-IID-Lighting~\citep{ke2025marigold}, which fine-tunes a pretrained diffusion backbone~\citep{rombach2022SD} on Hypersim~\citep{roberts2021hypersim} to produce albedo, diffuse shading, and a non-diffuse residual. We use it as a frozen extractor in both training and inference, and inherit its image-formation model as the bridge between real photographs and physically based renders.

\paragraph{Feed-forward dense prediction} has steadily replaced iterative pipelines across vision. VGGT~\citep{wang2025vggt} estimates cameras, depth, point maps, and tracks for hundreds of views in a single pass, displacing the bundle-adjustment loop of classical SfM~\citep{hartley2003mvgeometry}; DUSt3R~\citep{wang2024dust3r} and MASt3R~\citep{leroy2024mast3r} regress aligned pointmaps without explicit matching; RayZer~\citep{jiang2025rayzer} and LVSM~\citep{jin2024lvsm} synthesize novel views without an intermediate 3D representation. A generic transformer trunk, scale, and task-specific supervision suffice to match or surpass much of the task-specific machinery they replace. Single-image relighting has predominantly remained in the diffusion paradigm~\citep{liang2025diffusionrenderer, he2025unirelight, zhang2025iclight, zeng2024rgbx, fang2025vrgbx}, whose iterative sampling is slow and whose generative prior can drift from the input image -- a serious problem in relighting, where most pixels should remain photometrically faithful to the source. Even within diffusion, recent work pursues single-step formulations explicitly motivated by inference latency~\citep{sun2025ouroboros, serrano2026synclight}. We adopt the feed-forward recipe from the outset, with a per-pixel modulation parameterization that preserves source identity by construction.
\section{Method}
\label{sec:method}

We present \textsc{PIXLRelight}, a feed-forward transformer that relights a single input image to match a user-specified target lighting condition. The target condition is supplied as an intrinsic decomposition of the desired appearance, which serves as a unified conditioning interface for both training and inference. We define the problem in~\cref{sec:method:problem}, describe the architecture in~\cref{sec:method:arch}, and detail training and inference in~\cref{sec:method:train,sec:method:inference}.

\subsection{Problem definition}
\label{sec:method:problem}

Let $I_S \in [0,1]^{3 \times H \times W}$ be a source RGB image of a static scene under lighting $L_S$, and let $I_T \in [0,1]^{3 \times H \times W}$ be the same scene under a different target lighting $L_T$. We adopt the intrinsic image-formation model followed by~\citet{ke2025marigold},
\begin{equation}
  I \;=\; A \odot S \;+\; R,
  \label{eq:image_formation}
\end{equation}
where $A \in [0,1]^{3 \times H \times W}$ is the diffuse albedo, $S \in \mathbb{R}_{\geq 0}^{3 \times H \times W}$ is the diffuse shading, $R \in \mathbb{R}^{3 \times H \times W}$ is the non-diffuse residual capturing specular highlights, transparency, and other non-Lambertian effects, and $\odot$ denotes element-wise multiplication. The triplet $(A_T, S_T, R_T)$ thus fully encodes the target appearance under $L_T$: the albedo is shared with the source by construction, while $S_T$ and $R_T$ together carry every change induced by the target lighting.

We represent the target lighting condition through the channel-wise concatenation of the three target intrinsic maps:
\begin{equation}
  C_T \;=\; [\,A_T \,;\, S_T \,;\, R_T\,] \;\in\; \mathbb{R}^{9 \times H \times W}.
\end{equation}
We seek a function $f_\theta$ such that
\begin{equation}
  \hat{I}_T \;=\; f_\theta(I_S, C_T) \;\approx\; I_T .
\end{equation}
$C_T$ is the only carrier of lighting information seen by the model, and the same interface is used regardless of how $C_T$ is produced: from a real photograph at training (\cref{sec:method:train}), and from a physically based render at inference (\cref{sec:method:inference}).

\subsection{Architecture}
\label{sec:method:arch}

\begin{figure}[t]
  \centering
  \includegraphics[width=\linewidth]{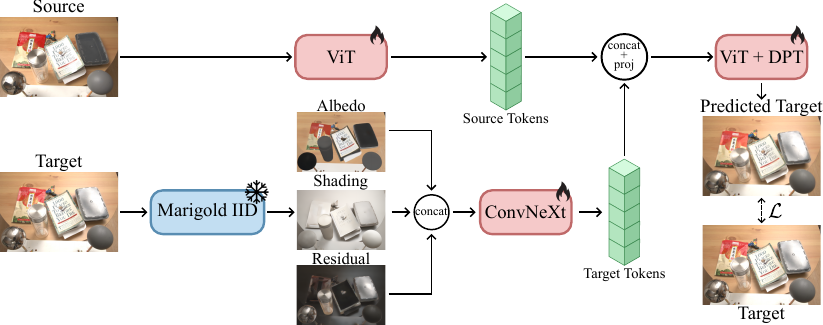}
  \caption{\textbf{Training pipeline.} The source image is patchified by a ViT branch and the channel-wise concatenated target intrinsics -- extracted from the target image by a frozen Marigold-IID-Lighting model -- are patchified by a ConvNeXt branch. The two token grids are fused per spatial location, projected to a common dimension, and processed by a self-attention transformer trunk. A DPT head reads out intermediate trunk features and predicts a per-pixel affine modulation of the source. Training is supervised end-to-end against the target image with pixel and perceptual losses.}
  \label{fig:training}
\end{figure}

\textsc{PIXLRelight} is a transformer that consumes the source image and the target intrinsics jointly and produces a relit RGB image. Its design follows recent feed-forward transformer architectures for dense prediction~\citep{wang2025vggt, jiang2025rayzer, lin2025depthanything3, jin2024lvsm}: a pair of asymmetric encoders feeds a shared transformer trunk, whose intermediate features are read out by a DPT head~\citep{ranftl2021dpt}. An overview is shown in~\cref{fig:training}.

\paragraph{Asymmetric Feature Encoders.}
The two inputs have very different spatial statistics, and we tokenize them accordingly. The source image $I_S$, dominated by high-frequency content, is patchified by a Vision Transformer (ViT)~\citep{vaswani2017attention, dosovitskiy2020vit}; the intrinsic stack $C_T$, dominated by smooth, lower-frequency structure, is patchified by a ConvNeXt~\citep{liu2022convnext, woo2023convnextv2}, whose convolutional inductive bias suits smoother inputs. Both branches use a patch size of $p$ and produce token grids of size $(H/p) \times (W/p)$ with embedding dimension $d$. We then perform per-location fusion: at each spatial position, we concatenate the source and conditioning tokens channel-wise and project to dimension $d$ with a small MLP, yielding a single fused token grid that is the input to the transformer trunk. Per-location fusion preserves the spatial layout of the conditioning and halves the sequence length compared with sequence-level concatenation.

\paragraph{Transformer Trunk.}
The fused token sequence is processed by a stack of $L$ self-attention transformer blocks. We prepend a small set of learnable register tokens~\citep{darcet2023registertokens} to provide a global communication channel, and apply two-dimensional Rotary Positional Embeddings~\citep{su2024rope} to the patch tokens to encode their spatial layout. Following DPT-style dense prediction~\citep{ranftl2021dpt, yang2024depthanything2}, we expose the outputs of four intermediate blocks for multi-scale fusion in the head.

\paragraph{Modulation Head.}
The four intermediate token streams are first converted to a single dense feature map $F \in \mathbb{R}^{C'\times H \times W}$ with a DPT layer~\citep{ranftl2021dpt}, then mapped with a $1{\times}1$ convolution to a six-channel output. Rather than regressing the relit RGB image directly, we parameterize the output as a per-pixel affine modulation of the source: most of the high-frequency content of $\hat{I}_T$ is already present in $I_S$, and asking the network to reproduce it from scratch wastes capacity that should be spent on transferring lighting. We split the six channels into a gain map $g \in \mathbb{R}^{3 \times H \times W}$ and a bias map $b \in \mathbb{R}^{3 \times H \times W}$, and form the prediction as
\begin{equation}
  \hat{I}_T \;=\; \mathrm{clip}\!\left( (1 + g) \odot I_S + b,\; 0,\; 1 \right) .
  \label{eq:modulation}
\end{equation}
The output convolution is initialized to zero, so $g \equiv 0$ and $b \equiv 0$ at initialization and the network outputs $\hat{I}_T = I_S$ exactly. The model thus starts from an identity prior and learns only the residual lighting transformation. Gradients flow through the clip wherever the prediction lies inside $[0,1]$, which holds at initialization and continues to hold during training.

\subsection{Training}
\label{sec:method:train}

\paragraph{Training Data.}
We train on paired captures of static scenes under varying illumination, combining three datasets: the MIT Multi-Illumination Images in the Wild dataset (MIIW)~\citep{miiw}, with 985 scenes captured under 25 different artificial flash conditions; BigTime~\citep{bigtime}, with 212 time-lapse scenes captured under varying natural illumination; and VIDIT~\citep{vidit}, with 300 synthetic Unreal Engine scenes rendered under 40 lighting conditions formed by 5 color temperatures and 8 light directions. Together they span controlled artificial and uncontrolled natural lighting, real and synthetic captures, and indoor and outdoor scenes. Although smaller than the unpaired photo collections used by self-supervised relighting methods, these datasets provide paired multi-illumination supervision, which directly trains the photometric transfer we need without requiring per-image rendering optimizations~\citep{careaga2025physically}. For each batch, we randomly sample two images of the same scene and randomly assign them the roles of source and target, yielding dense supervision for arbitrary directional lighting changes between any two captured conditions. The target intrinsics $C_T$ are produced on the fly by a frozen Marigold-IID-Lighting model~\citep{ke2025marigold} run for a single denoising step.

\paragraph{Training Loss.}
We supervise the predicted relit image directly against the ground-truth target image with a photometric loss,
\begin{equation}
  \mathcal{L}(\hat{I}_T, I_T) \;=\; \lVert \hat{I}_T - I_T \rVert_1
   \;+\; \lambda \, \mathcal{L}_\mathrm{perc}(\hat{I}_T, I_T),
  \label{eq:loss}
\end{equation}
where $\mathcal{L}_\mathrm{perc}$ is a VGG-based perceptual loss~\citep{johnson2016perceptualloss, zhang2018perceptual} and $\lambda$ balances the two terms. The gain and bias maps are not supervised directly; they emerge from end-to-end training with the only objective being to match the target image.

\paragraph{Implementation Details.}
The RGB encoder is a ViT-Large ($24$ blocks, $d{=}1024$, $16$ heads); the intrinsics encoder is a ConvNeXt-Base whose final-stage features are projected to dimension $d{=}1024$. The transformer trunk consists of $L{=}24$ self-attention blocks with $d{=}1024$, $16$ heads, and $8$ register tokens; we feed the outputs of blocks $\{4, 11, 17, 23\}$ to the DPT head. All branches use patch size $p{=}16$. The model has approximately $640$M parameters. We train with AdamW~\citep{loshchilov2017adamw} for $200$K iterations using a cosine learning-rate schedule peaking at $5{\times}10^{-5}$ after $2.5$K warmup steps, with $\lambda{=}0.2$ and gradients clipped at $1.0$. Training uses bfloat16 mixed precision and gradient checkpointing on two H200 GPUs and takes approximately four days. Input images are resized to $512$-pixel longer side with random aspect ratio in $[0.33, 1.0]$ and random horizontal flips; we deliberately avoid photometric augmentations on the source and target, which would corrupt the lighting signal the model is supervised to learn. We do, however, apply corruption augmentations to the conditioning $C_T$ to simulate the artifacts produced by single-image geometry and material reconstruction at inference (\cref{sec:appendix:augmentation}). Full hyperparameters are in~\cref{sec:appendix:training}.

\subsection{Inference}
\label{sec:method:inference}

\begin{figure}[t]
  \centering
  \includegraphics[width=\linewidth]{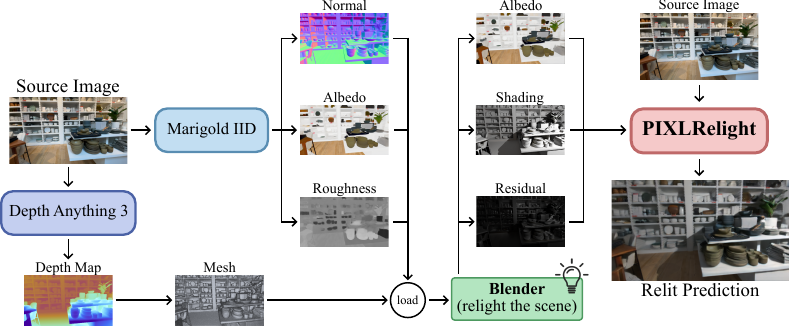}
  \caption{\textbf{Inference pipeline.} Given a single input image, geometry is recovered by Depth Anything 3 and unprojected to a triangle mesh, and materials are recovered by Marigold-IID-Appearance. The textured mesh is loaded into Blender, where the user authors the desired illumination; Blender Cycles then renders the scene and produces the target intrinsic maps $C_T$. \textsc{PIXLRelight} takes as input the original image together with $C_T$ and produces the final relit prediction.}
  \label{fig:inference}
\end{figure}

At inference, the user provides a single image $I_S$ and wishes to relight it under a freely chosen target lighting condition. Our conditioning interface requires a target intrinsic decomposition $C_T$ that the user cannot directly author, so we bridge this gap with a physically based renderer (\cref{fig:inference}):

\newpage
\begin{enumerate}\itemsep0pt
  \item \textbf{Geometry.} A metric depth map for $I_S$ is estimated with Depth Anything~3~\citep{lin2025depthanything3} and unprojected into a triangle mesh.
  \item \textbf{Materials.} Marigold-IID-Appearance~\citep{ke2025marigold} extracts pixel-aligned albedo, surface normal, and roughness maps from $I_S$.
  \item \textbf{User edit.} The mesh and materials are imported into Blender, where the user freely authors the desired illumination using arbitrary combinations of area lights, environment maps, sun lamps, and emissive geometry.
  \item \textbf{Conditioning.} The three target intrinsic buffers $C_T = [A_T \,;\, S_T \,;\, R_T]$ are composed directly from Blender's Cycles render passes following the image-formation model of~\cref{eq:image_formation}; we provide the exact processing formulas in~\cref{sec:appendix:blender}.
  \item \textbf{Relit prediction.} The original input image $I_S$ together with $C_T$ is passed to \textsc{PIXLRelight}, which produces the final relit prediction $\hat{I}_T$ in a single forward pass.
\end{enumerate}

The Blender render of the scene is never shown to the model. Single-image geometry and material estimates from off-the-shelf tools are coarse, and the rendered RGB inherits these errors. The intrinsic buffers $C_T$ are still a valid lighting specification, because errors in geometry or materials corrupt $C_T$ locally, at the affected pixels, rather than propagating globally through the rendered image. This locality, combined with the corruption augmentations of~\cref{sec:appendix:augmentation}, allows \textsc{PIXLRelight} to be conditioned on coarse intrinsic buffers and still produce photorealistic relightings.
\section{Experiments}
\label{sec:experiments}

\subsection{Quantitative comparison}
\label{sec:exp:quantitative}

\paragraph{Baselines.}
We compare against five recent baselines, grouped by the lighting cue they consume. DiffusionRenderer~\citep{liang2025diffusionrenderer} and UniRelight~\citep{he2025unirelight} consume an HDR environment map; since neither accepts an arbitrary target image, we estimate the target environment from the ground-truth target with DiffusionLight-Turbo~\citep{chinchuthakun2026diffusionlightturbo}. RGBX~\citep{zeng2024rgbx}, Ouroboros~\citep{sun2025ouroboros}, and V-RGBX~\citep{fang2025vrgbx} consume a target diffuse-shading map alongside source G-buffers; we drive each with its own inverse-rendering stage, taking source G-buffers from the source image and target shading from the target image. All baselines use their official released checkpoints. We do not compare against~\citet{careaga2025physically}, the closest prior work, because no code or model has been released.

\paragraph{Metrics.}
We report PSNR, SSIM~\citep{wang2004ssim}, and LPIPS~\citep{zhang2018perceptual} at native target resolution. Following~\citep{kocsis2024iid, jin2024neuralgaffer, liang2025diffusionrenderer, he2025unirelight}, we apply a per-image, per-channel least-squares scale correction to absorb global exposure ambiguity, uniformly across every method including ours. Inference times are forward-pass wall-clock on an NVIDIA RTX A6000, averaged over five runs after two warm-ups.

\paragraph{Datasets.}
We evaluate on the official test split of MIT Multi-Illumination Images in the Wild~\citep{miiw}, comprising 30 indoor scenes. We additionally collect a small held-out set of six indoor scenes captured on a stationary tripod under two everyday lighting conditions each, evaluated in both directions for twelve source--target pairs. Neither benchmark overlaps with our training data, and the held-out set probes a lighting distribution distinct from MIIW's controlled flashes.

\begin{figure}[t]
  \centering
  \includegraphics[width=\linewidth]{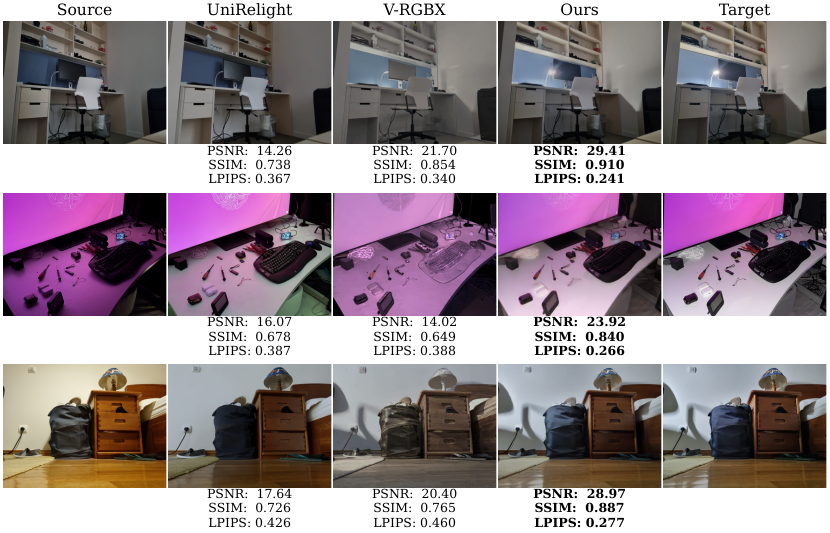}
  \caption{\textbf{Held-out tripod captures: paired source--target relighting.} Three representative pairs from a held-out set of six indoor scenes captured under two lighting conditions each. For visual clarity we display the most recent baseline per conditioning group: UniRelight (environment-map) and V-RGBX (shading). Per-image PSNR/SSIM/LPIPS are shown below each prediction; \textbf{bold} marks the best method per image. \textsc{PIXLRelight} is best on every metric in every scene.}
  \label{fig:examples_qualitative}
\end{figure}

\paragraph{Results.}
\Cref{tab:quant_eval} reports MIIW test-split results. \textsc{PIXLRelight} outperforms every baseline by a wide margin, exceeding the next-best result on each metric by $9.8$~dB PSNR and $0.130$ SSIM (over Ouroboros) and $0.243$ LPIPS (over RGBX), and runs in $0.09$~s per image -- at least an order of magnitude faster than every baseline. The margin holds on the held-out tripod set (\cref{fig:examples_qualitative}), where \textsc{PIXLRelight} is best on every metric in every scene, with per-scene PSNR gaps of $8$--$9$~dB. The two baseline groups fail in distinct ways. UniRelight (environment-map cue) cannot represent the spatially varying near-field sources that dominate indoor scenes: in row~1 it misses the highlight cast by the desk lamp, and in row~3 the directional shadow behind the backpack is absent. V-RGBX (shading cue) inverse-renders source and target into G-buffers and shading, then re-renders both with a forward diffusion model; this chain fails on both ends -- it neither preserves the source nor transfers the target lighting, missing the desk light in row~1, retaining the source's strong magenta cast in row~2, and washing out the lamp scene in row~3. \textsc{PIXLRelight} avoids both failure modes: the full intrinsic stack carries the target lighting while the source RGB carries photographic detail. More comparisons in~\cref{sec:appendix:experiments_miiw,sec:appendix:experiments_examples}.

\begin{table}[!ht]
\centering
\begin{threeparttable}
\caption{\textbf{Quantitative evaluation on the MIIW test split~\citep{miiw}.} Methods are grouped by the target-lighting cue they consume. All metrics use a per-image, per-channel least-squares scale correction applied uniformly to every method, including ours~\citep{kocsis2024iid, jin2024neuralgaffer, liang2025diffusionrenderer, he2025unirelight}. Inference times are wall-clock times of one relighting forward pass on an NVIDIA RTX A6000, averaged over 5 runs after 2 warm-ups; intrinsic estimation, which differs across methods, is excluded. \textbf{Bold}: best; \underline{underline}: second-best.}
\label{tab:quant_eval}
\small
\begin{tabular}{lccccc}
\toprule
\textbf{Method} & \textbf{Lighting cue} & PSNR $\uparrow$ & SSIM $\uparrow$ & LPIPS $\downarrow$ & Time (s) $\downarrow$ \\
\midrule
DiffusionRenderer~\citep{liang2025diffusionrenderer} & \multirow{2}{*}{Environment map} & 15.02 & 0.663 & 0.461 & 3.37 \\
UniRelight~\citep{he2025unirelight} & & 16.71 & 0.672 & 0.444 & 58.58 \\
\noalign{\smallskip}\hdashline\noalign{\smallskip}
RGBX~\citep{zeng2024rgbx} & \multirow{3}{*}{Diffuse shading} & 16.20 & 0.690 & \underline{0.438} & 13.98 \\
Ouroboros~\citep{sun2025ouroboros} & & \underline{19.38} & \underline{0.763} & 0.463 & \underline{0.62} \\
V-RGBX~\citep{fang2025vrgbx} & & 17.79 & 0.725 & 0.467 & 841.13$^\dagger$ \\
\midrule
\textbf{Ours} & Diffuse shading & \textbf{29.18} & \textbf{0.893} & \textbf{0.195} & \textbf{0.09} \\
\bottomrule
\end{tabular}
\begin{tablenotes}\footnotesize
\item[$\dagger$] V-RGBX produces noise on a single-frame input; we replicate the source $49$ times to reach the model's minimum supported sequence length and report one full forward pass over that sequence.
\end{tablenotes}
\end{threeparttable}
\end{table}

\subsection{Controllable relighting from authored illumination}
\label{sec:experiments:controllable}

The previous evaluations test how faithfully each method transfers a captured target lighting, but cannot test controllability. We now evaluate relighting under physically authored target illumination, supplied through the same intrinsic interface used at training but produced by a path tracer at inference.

\paragraph{Protocol.}
We automate the inference pipeline of~\cref{sec:method:inference} into a Blender script: given an input photograph, it recovers a textured mesh (Depth Anything~3~\citep{lin2025depthanything3} for geometry, Marigold-IID-Appearance~\citep{ke2025marigold} for albedo/normals/roughness), inserts one of five preset lighting setups (cool side flash, warm overhead flash, dim overhead spot, soft frontal sun, warm interior sun), and renders the scene in linear HDR, exporting the Cycles passes needed to compose $C_T$ via~\cref{eq:image_formation} (formulas in~\cref{sec:appendix:blender}). \textsc{PIXLRelight} consumes $C_T$ alongside the source image; shading-conditioned baselines consume only the diffuse-shading channel of $C_T$ together with their own inverse-rendered source G-buffers. We exclude DiffusionRenderer and UniRelight, which require an HDR environment map. We apply this pipeline to twenty in-the-wild images from DL3DV~\citep{dl3dv} under all five setups; with no ground-truth relit photograph available, the evaluation is qualitative. For display, sRGB method outputs are shown as produced; the linear-HDR path traced reference is tonemapped to sRGB via Reinhard auto-exposure (key=0.18).

\begin{figure}[t]
  \centering
  \includegraphics[width=\linewidth]{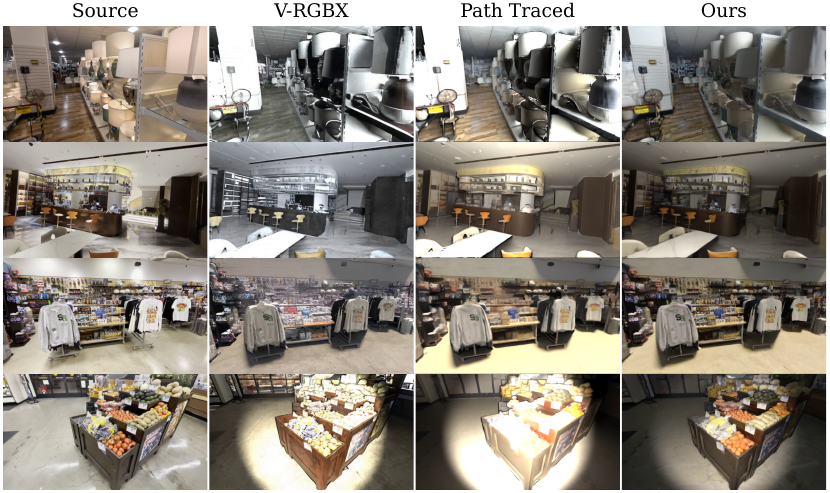}
  \caption{\textbf{Relighting from path-traced illumination on DL3DV scenes~\citep{dl3dv}.} Each row shows a source image, V-RGBX (the most recent shading-conditioned baseline), Blender's full RGB render of the reconstructed scene under the authored lighting (Path Traced), and \textsc{PIXLRelight}. V-RGBX produces plausible relightings but drifts from the source. \textsc{PIXLRelight} transfers the authored lighting while preserving the source's photographic detail.}
  \label{fig:dl3dv_qualitative}
\end{figure}

\paragraph{Results.}
\Cref{fig:dl3dv_qualitative} shows five DL3DV scenes relit under Blender-authored illumination, alongside the path-traced render of each reconstruction. Without ground truth, identity preservation -- not plausibility alone -- separates the methods. V-RGBX, the most recent shading-conditioned baseline, drifts from the source: it desaturates the scene in rows~1, 2, and 5, and overshoots the requested spotlight in row~4. \textsc{PIXLRelight} reproduces the authored lighting in each row -- side-lit shadows in the lamp store (row~1), warm interior glow in the bar (row~2) and empty room (row~5), warm overhead falloff in the gift shop (row~3), and a focused spotlight on the fruit stand (row~4) -- while leaving the underlying scene unchanged. The Path Traced column reveals a limitation common to all single-image PBR pipelines, including the closest prior work~\citep{careaga2025physically}: the underlying 3D reconstruction is coarse and the recovered materials are imperfect, so the rendered RGB visibly drifts from the source. \textsc{PIXLRelight} is robust to this drift because the source image and $C_T$ enter the model through separate branches: the source carries photographic content and $C_T$ carries the intrinsic components, so the model never has to disentangle the two from a single rendered RGB. Further results are in~\cref{sec:appendix:experiments_dl3dv}.

\subsection{Ablation studies}
\label{sec:experiments:ablation}

We ablate the two principal architectural choices of \textsc{PIXLRelight}: the fusion of source and intrinsic features inside the transformer trunk, and the modulation head. Both variants are trained from scratch under the protocol of~\cref{sec:method:train} and differ from the full model in exactly one component.~\Cref{tab:ablation} reports MIIW results. Removing the source ViT branch -- so the source enters the network only through the modulation head -- costs $3.36$~dB PSNR, $0.062$ SSIM, and $0.179$ LPIPS; without scene structure available to self-attention, predictions hew to the source illumination rather than transferring the requested condition (see~\cref{sec:appendix:ablation}). Replacing the modulation head with a direct-RGB regression costs $1.41$~dB PSNR, $0.060$ SSIM, and $0.164$ LPIPS; the loss is consistent but visually subtle, since direct regression still recovers the global lighting but must regenerate source-aligned texture from scratch. Both ablated variants still outperform every baseline in~\cref{tab:quant_eval}, indicating that the supervision regime -- direct training on paired multi-illumination captures -- drives most of our gains, with the architectural choices providing the rest.

\begin{table}[!h]
\centering
\caption{\textbf{Architectural ablations on the MIIW test split.} Both variants are trained from scratch and differ from the full model in exactly one component. Intrinsics-only trunk: the source ViT branch is removed; the source enters the network only through the modulation head. Direct regression head: the modulation of~\cref{eq:modulation} is replaced by a sigmoid-activated RGB regression. Both variants still beat every baseline in~\cref{tab:quant_eval}.}
\label{tab:ablation}
\small
\begin{tabular}{lccc}
\toprule
\textbf{Variant} & PSNR $\uparrow$ & SSIM $\uparrow$ & LPIPS $\downarrow$ \\
\midrule
Intrinsics-only trunk          & 25.82 & 0.831 & 0.374 \\
Direct regression head         & 27.77 & 0.833 & 0.359 \\
\midrule
\textbf{Ours}                  & \textbf{29.18} & \textbf{0.893} & \textbf{0.195} \\
\bottomrule
\end{tabular}
\end{table}
\section{Conclusion}
\label{sec:conclusion}

We present \textsc{PIXLRelight}, a feed-forward transformer that brings the physical lighting control of computer graphics to in-the-wild photographs. By separating \emph{what} the target lighting should be from \emph{how} it is applied to a real photograph, and bridging the two through a single intrinsic-decomposition interface, we train directly on paired multi-illumination photographs and accept arbitrary path-traced illumination at inference. \textsc{PIXLRelight} achieves state-of-the-art relighting quality in under a tenth of a second per image -- an order of magnitude faster than prior approaches -- enabling interactive lighting authoring on real photographs.

{\small
\bibliographystyle{ieeenat_fullname}
\bibliography{sections/11_references}
}

\newpage
\appendix
{\Large\bfseries Appendix}
\vspace{0.5em}

This appendix collects implementation details and additional results. \Cref{sec:appendix:training} lists the full training hyperparameters; \cref{sec:appendix:blender} describes how the target intrinsic conditioning $C_T$ is composed from Blender's Cycles render passes at inference; \cref{sec:appendix:augmentation} details the corruption augmentations applied to $C_T$ during training; and \cref{sec:appendix:limitations} discusses the limitations and failure modes of our pipeline. The remaining sections present additional qualitative results: an extended version of the banner figure (\cref{sec:appendix:experiments_banner}), additional scenes from the MIIW test split (\cref{sec:appendix:experiments_miiw}), further scenes from the held-out tripod set (\cref{sec:appendix:experiments_examples}), additional comparisons on relit DL3DV scenes (\cref{sec:appendix:experiments_dl3dv}), and qualitative ablations (\cref{sec:appendix:ablation}).

\section{Training details}
\label{sec:appendix:training}
We collect here the hyperparameters omitted from~\cref{sec:method:train}.~\Cref{tab:hyperparams} lists the optimization, data, architecture, and compute settings used to produce the results reported in the main paper. All hyperparameters were selected through small-scale runs and held fixed for the final training; we did not perform extensive sweeps.
\begin{table}[ht]
\centering
\caption{\textbf{Training hyperparameters for \textsc{PIXLRelight}.}}
\label{tab:hyperparams}
\small
\begin{tabular}{lll}
\toprule
\textbf{Group} & \textbf{Hyperparameter} & \textbf{Value} \\
\midrule
\multirow{10}{*}{Optimization}
 & Optimizer                     & AdamW~\citep{loshchilov2017adamw} \\
 & $(\beta_1, \beta_2)$          & $(0.9, 0.95)$ \\
 & Weight decay                  & $0.05$ \\
 & Peak learning rate            & $5\times10^{-5}$ \\
 & Final learning rate           & $1\times10^{-5}$ \\
 & Schedule                      & cosine, $2{,}500$ warmup steps \\
 & Iterations                    & $200{,}000$ \\
 & Batch size (per GPU)          & $42$ \\
 & Effective batch size          & $84$ ($2$ GPUs) \\
 & Gradient clipping (max norm)  & $1.0$ \\
 & Mixed precision               & bfloat16 \\
\midrule
\multirow{6}{*}{Data}
 & Datasets                      & MIIW~\citep{miiw}, BigTime~\citep{bigtime}, VIDIT~\citep{vidit} \\
 & Longer-side resolution        & $512$ \\
 & Random aspect ratio           & $[0.33, 1.0]$ \\
 & Random horizontal flip        & yes \\
 & Photometric augmentations     & none on the source/target (would corrupt the lighting signal) \\
 & Conditioning augmentations    & corruption pipeline on $C_T$ (\cref{sec:appendix:augmentation}) \\
\midrule
\multirow{8}{*}{Architecture}
 & Source encoder                & ViT-Large ($24$ blocks, $d{=}1024$, $16$ heads) \\
 & Intrinsics encoder            & ConvNeXt-Base, projected to $d{=}1024$ \\
 & Trunk depth                   & $L=24$ self-attention blocks \\
 & Trunk width                   & $d=1024$, $16$ heads \\
 & Register tokens               & $8$ \\
 & DPT readout blocks            & $\{4, 11, 17, 23\}$ \\
 & Patch size                    & $p=16$ \\
 & RoPE base frequency           & $100.0$ \\
 & Total parameters              & $\approx 640$M \\
\midrule
\multirow{2}{*}{Loss}
 & Pixel loss                    & $\ell_1$ \\
 & Perceptual loss weight $\lambda$ & $0.2$ \\
\midrule
\multirow{2}{*}{Compute}
 & Hardware                      & $2{\times}$NVIDIA H200 \\
 & Wall-clock time               & $\approx 4$ days \\
\bottomrule
\end{tabular}
\end{table}

\newpage
\section{Composing target intrinsics from Blender}
\label{sec:appendix:blender}
To produce the target intrinsic conditioning $C_T$ at inference time, we read Blender's Cycles render passes for the user-lit scene and compose them following the image-formation model of Marigold-IID-Lighting~\citep{ke2025marigold}. The albedo and diffuse shading are obtained directly from the diffuse passes:
\begin{align}
  A_T &= \mathrm{clip}(\mathit{diffuse\_color},\,0,\,1), \\
  S_T &= \max\!\left(\mathit{diffuse\_direct} + \mathit{diffuse\_indirect},\,0\right).
\end{align}
The non-diffuse residual $R_T$ aggregates every non-Lambertian light-transport contribution that Cycles exposes as a render pass -- glossy reflection, transmission (refraction and transparency), participating media (volume scattering), and self-emission:
\begin{equation}
\begin{aligned}
  R_T \;=\; \max\!\Big(\;
    & \mathit{glossy\_color} \odot
        \left( \mathit{glossy\_direct} + \mathit{glossy\_indirect} \right) \\
    {} +{} & \mathit{transmission\_color} \odot
        \left( \mathit{transmission\_direct} + \mathit{transmission\_indirect} \right) \\
    {} +{} & \left( \mathit{volume\_direct} + \mathit{volume\_indirect} \right) \\
    {} +{} & \mathit{emission}
  \,\;,\,0\Big),
\end{aligned}
\end{equation}
where the $\mathit{*\_color}$ terms are the per-material reflectance/transmittance passes, the $\mathit{*\_\{direct,indirect\}}$ terms are the corresponding direct- and indirect-lighting passes, $\mathit{volume\_\{direct,indirect\}}$ are the in-scattered radiance from participating media, $\mathit{emission}$ is the self-emission pass, and $\odot$ is element-wise multiplication. This decomposition exhausts the non-diffuse light-transport channels Cycles makes available, so any path-traced lighting effect not encoded in the diffuse passes -- specular highlights, refraction through transparent objects, glow from emissive geometry, or scattering through fog -- is captured by $R_T$.

The albedo $A_T$ lies in $[0,1]$ by construction, but $S_T$ and $R_T$ are HDR. To match the distribution that Marigold-IID-Lighting was trained on, we apply a joint $98^{\text{th}}$-percentile rescaling: we compute $\tau = \max\!\left(\mathrm{p98}(S_T),\,\mathrm{p98}(R_T),\,\varepsilon\right)$ and replace $S_T \leftarrow \mathrm{clip}(S_T, 0, \tau)/\tau$ and $R_T \leftarrow \mathrm{clip}(R_T, 0, \tau)/\tau$. Sharing the cutoff $\tau$ across both channels preserves their relative magnitudes, which encodes the diffuse-to-specular balance of the target lighting.

\section{Conditioning augmentations}
\label{sec:appendix:augmentation}
The intrinsic conditioning $C_T$ seen at inference is composed from Blender render passes of a coarse, single-image reconstruction. It carries artifacts that never appear in $C_T$ at training, where it is extracted from a real photograph: missing-geometry holes, silhouette cracks at depth discontinuities, render speckle, denoiser blur, posterized banding, and per-channel exposure or color shifts inherited from the upstream estimators. Training only on clean, photograph-derived $C_T$ would let the model overfit to its smooth statistics and degrade at inference. We therefore apply a stochastic corruption pipeline to $C_T$ during training, designed to resemble these artifacts.

The pipeline is applied per sample, on GPU, after Marigold-IID-Lighting and before the conditioning encoder. It consists of eight independently gated augmentations grouped into four families:
\begin{itemize}\itemsep0pt
  \item \textbf{Photometric.} Per-channel multiplicative \emph{color cast} and additive bias; per-channel \emph{gamma}.
  \item \textbf{Structural.} \emph{Holes}: a low-resolution bilinearly-upsampled noise mask, optionally biased toward Sobel edges, replaces a sample-specific top fraction of pixels with the channel minimum plus a small noise floor. \emph{Edge cracks}: a quantile threshold on the Sobel edge map produces a narrow silhouette mask which is dilated and used to multiplicatively darken those pixels.
  \item \textbf{Noise.} Per-pixel \emph{salt-and-pepper} replacement and additive \emph{Gaussian} noise.
  \item \textbf{Frequency.} Separable \emph{Gaussian blur} (denoiser-style smoothing) and \emph{posterization} into a sample-specific number of levels.
\end{itemize}
Augmentations are applied in the order photometric $\to$ structural $\to$ noise $\to$ blur, so that downstream noise stacks on top of the perturbed tonal range and the introduced structural defects. A global Bernoulli gate $p_\mathrm{apply}=0.7$ wraps the entire pipeline, so $30\%$ of samples remain strictly clean; among the rest, each augmentation fires independently with its own per-sample probability, and the output is finally clipped to the input value range. The full set of probabilities and strength ranges is listed in~\cref{tab:augmentations}; values are deliberately mild and designed to mimic the renderer's failure modes rather than destroy the lighting signal.
\begin{table}[ht]
\centering
\caption{\textbf{Conditioning augmentation parameters.} The pipeline as a whole fires with probability $p_\mathrm{apply}$; conditional on firing, each augmentation fires independently per sample with the listed probability and its strength is sampled uniformly from the range. Photometric augmentations operate per intrinsic group (albedo, shading, residual).}
\label{tab:augmentations}
\small
\begin{tabular}{lll}
\toprule
\textbf{Augmentation} & \textbf{Probability} & \textbf{Strength range} \\
\midrule
Pipeline applied ($p_\mathrm{apply}$) & $0.70$ & --- \\
\midrule
Color cast (per-channel scale)        & $0.50$ & scale $\in [0.85, 1.15]$, bias $\in [-0.06, 0.06]$ \\
Gamma                                 & $0.30$ & $\gamma \in [0.75, 1.35]$ \\
\midrule
Holes (edge-biased)                   & $0.50$ & affected fraction $\in [0.005, 0.040]$ \\
Edge cracks (silhouette darkening)    & $0.50$ & quantile $\in [0.92, 0.99]$, strength $\in [0.4, 1.0]$ \\
\midrule
Salt-and-pepper                       & $0.25$ & per-pixel fraction $\in [0, 0.003]$ \\
Gaussian noise                        & $0.50$ & $\sigma \in [0, 0.04]$ \\
\midrule
Gaussian blur                         & $0.25$ & $\sigma \in [0.4, 1.2]$ \\
Posterization                         & $0.25$ & levels $\in [12, 64]$ \\
\bottomrule
\end{tabular}
\end{table}

\newpage
\section{Limitations and failure cases}
\label{sec:appendix:limitations}
\begin{figure}[!ht]
    \centering
    \includegraphics[width=\linewidth]{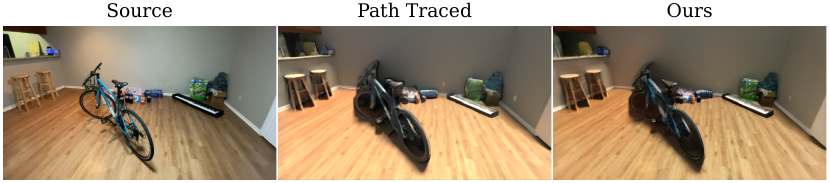}
    \caption{\textbf{Failure case from upstream reconstruction errors.} A DL3DV scene where the single-image depth estimator collapses the bicycle into the floor. The path-traced render carries this error into $C_T$, and \textsc{PIXLRelight} inherits it in the relit output.}
    \label{fig:limitations}
\end{figure}

\textsc{PIXLRelight} inherits the failure modes of its frozen dependencies. At training, the target intrinsics are produced by Marigold-IID-Lighting~\citep{ke2025marigold}; systematic errors in its decomposition -- such as baking a cast shadow into albedo, or attributing a colored highlight to diffuse rather than non-diffuse shading -- bias the supervisory signal. At inference, the same decomposer is applied to a Blender render, and the upstream geometry (Depth Anything~3~\citep{lin2025depthanything3}) and material (Marigold-IID-Appearance~\citep{ke2025marigold}) estimators add their own errors. The corruption augmentations of~\cref{sec:appendix:augmentation} make the model robust to local artifacts, but global reconstruction failures still propagate: when entire objects are mis-localized or fused with the background, the resulting $C_T$ no longer specifies the user's intended lighting on the original geometry. \Cref{fig:limitations} shows one such case on a DL3DV scene where the depth estimator collapses a bicycle into the floor; the path-traced render carries this error into $C_T$, and \textsc{PIXLRelight} -- which has no direct view of the original geometry beyond the source RGB -- inherits it in the relit output. We expect future improvements in feed-forward geometry, materials, and intrinsic decomposition to translate directly into better authoring fidelity, without retraining the relighting network.

A second limitation is the relatively small training corpus: $985 + 212 + 300 \approx 1{,}500$ scenes from MIIW, BigTime, and VIDIT is still two orders of magnitude smaller than the unpaired photo collections used by self-supervised relighting methods. Although our quantitative margin and the held-out tripod evaluation indicate that the intrinsic-conditioning interface generalizes beyond the training distribution, scaling paired multi-illumination supervision -- whether through new captures, simulated multi-illumination renders, or synthetic-to-real adaptation -- is a natural avenue for further gains.

\clearpage
\section{Multi-light authored relighting on a single scene}
\label{sec:appendix:experiments_banner}
\begin{figure}[!ht]
    \centering
    \includegraphics[width=0.85\linewidth]{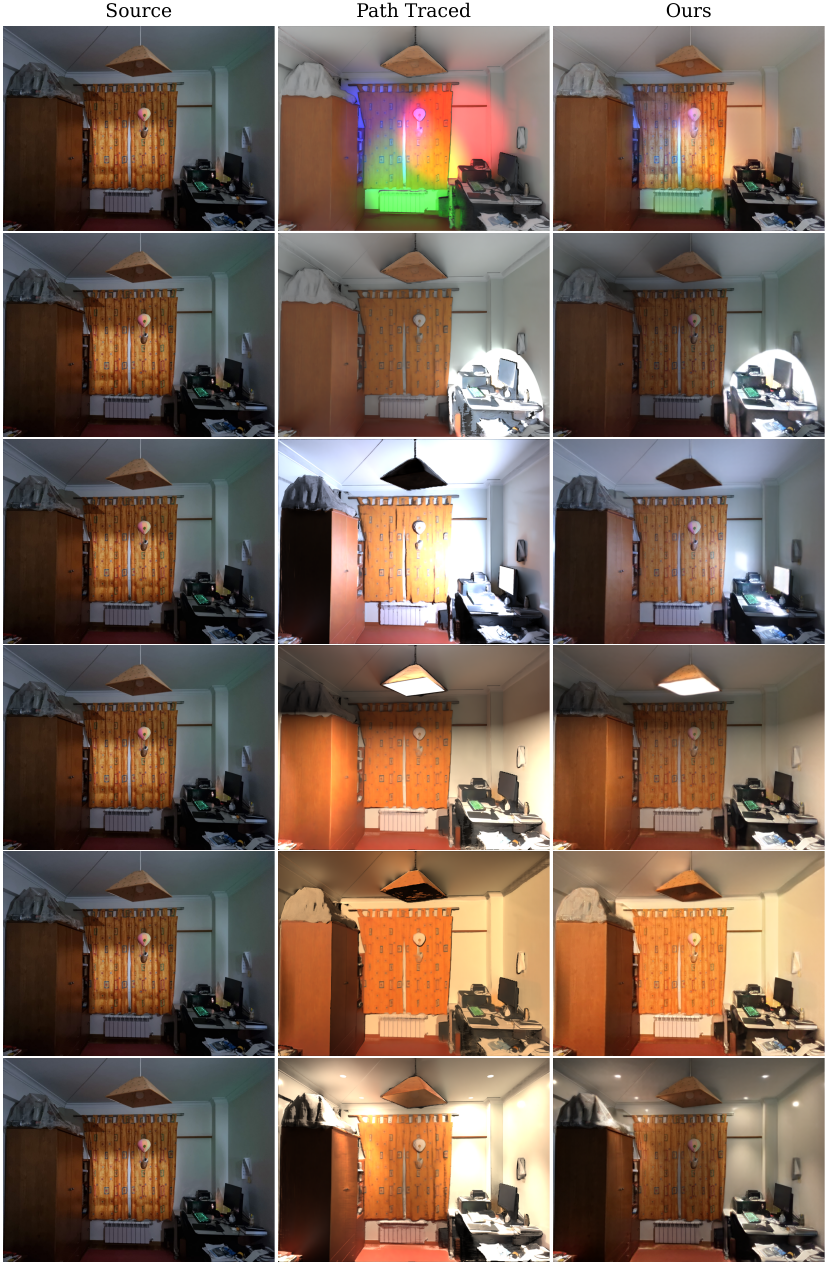}
    \caption{\textbf{Banner figure expansion.} A single source image (left) is relit by \textsc{PIXLRelight} (right) under six different Blender-authored illuminations. The middle column shows the corresponding path-traced render of the reconstructed scene. \textsc{PIXLRelight} consumes only the intrinsic buffers derived from these renders, together with the source image, and produces a sharper and more photorealistic relighting that retains the source's photographic detail while transferring the authored lighting.}
    \label{fig:banner_appendix}
\end{figure}

\clearpage
\section{Additional qualitative results on MIIW}
\label{sec:appendix:experiments_miiw}
\begin{figure}[!ht]
    \centering
    \includegraphics[width=\linewidth]{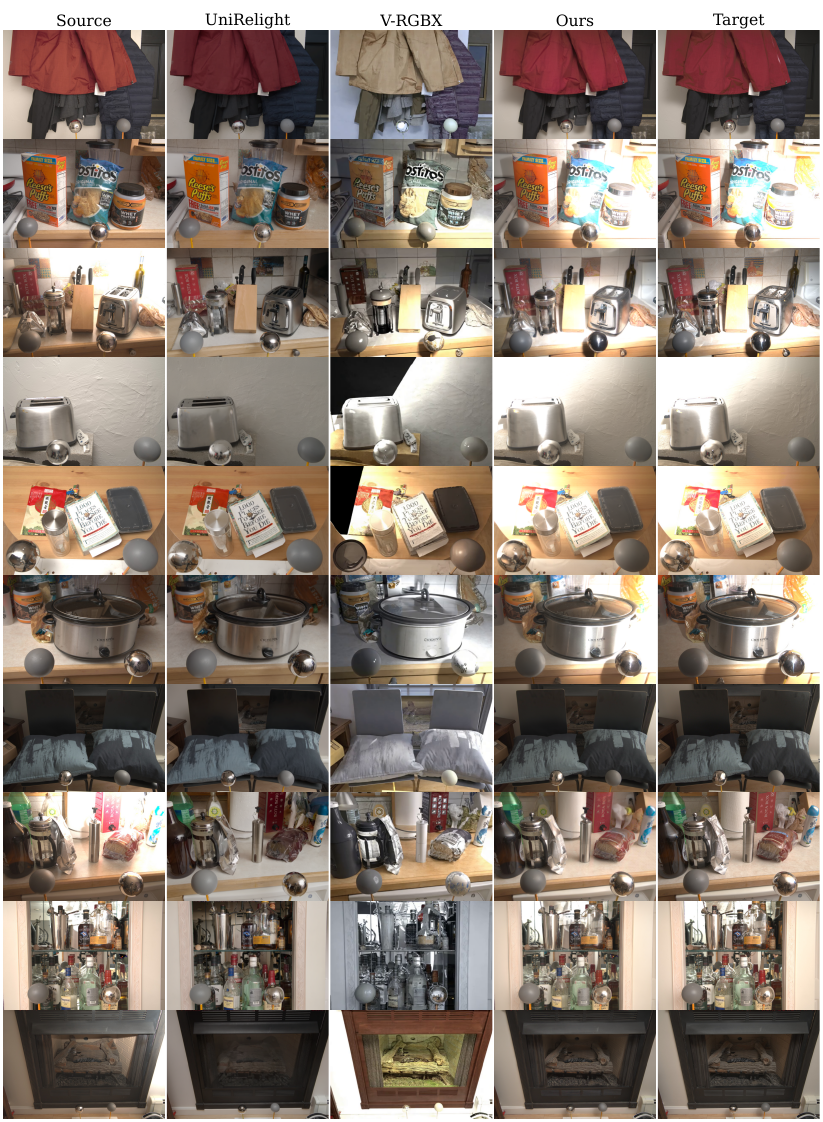}
    \caption{\textbf{Additional qualitative comparisons on the MIIW test split~\citep{miiw}.} Each row shows a single source--target pair, with predictions from the most recent baseline of each conditioning group (UniRelight for environment-map methods; V-RGBX for shading-conditioned methods, see~\cref{tab:quant_eval}) and \textsc{PIXLRelight}. Across all twelve scenes, \textsc{PIXLRelight} retains source detail by construction and transfers only the lighting change implied by the conditioning intrinsics, including specular highlights on chrome spheres and shading gradients on diffuse surfaces.}
    \label{fig:miiw_qualitative}
\end{figure}

\clearpage
\section{Additional held-out tripod captures}
\label{sec:appendix:experiments_examples}
\begin{figure}[!ht]
    \centering
    \includegraphics[width=\linewidth]{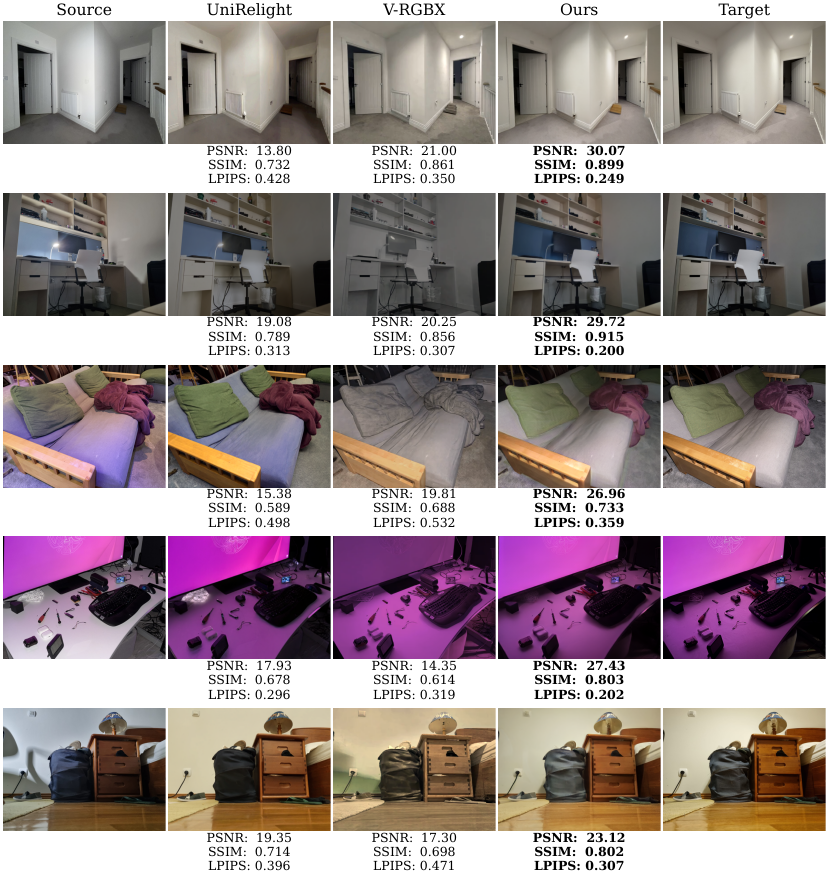}
    \caption{\textbf{Additional held-out tripod captures.} Further source--target pairs from the held-out set, beyond the three shown in~\cref{fig:examples_qualitative}. Columns: source, UniRelight (environment-map baseline), V-RGBX (shading baseline), \textsc{PIXLRelight}, and target. \textsc{PIXLRelight} is best on every metric in every scene.}
    \label{fig:examples_appendix}
\end{figure}

\clearpage
\section{Additional qualitative results on DL3DV scenes}
\label{sec:appendix:experiments_dl3dv}
\begin{figure}[!ht]
    \centering
    \includegraphics[width=\linewidth]{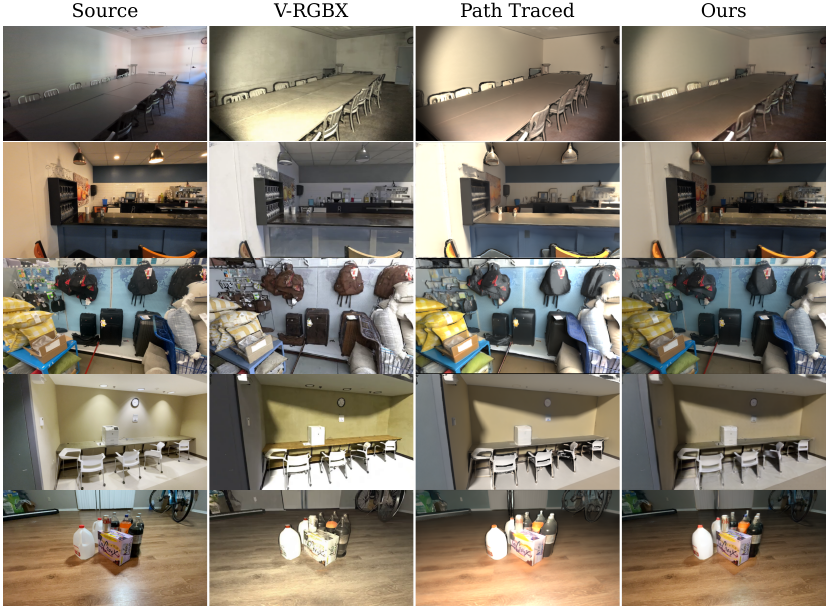}
    \caption{\textbf{Additional relighting comparisons on DL3DV scenes~\citep{dl3dv}.} Each row: a single source image, the most recent shading-conditioned baseline (V-RGBX), Blender's full RGB render of the reconstructed scene under the authored lighting (Path Traced), and \textsc{PIXLRelight}. \textsc{PIXLRelight} transfers the authored lighting while preserving the source's photographic detail.}
    \label{fig:dl3dv_qualitative_additional1}
\end{figure}

\begin{figure}[!ht]
    \centering
    \includegraphics[width=\linewidth]{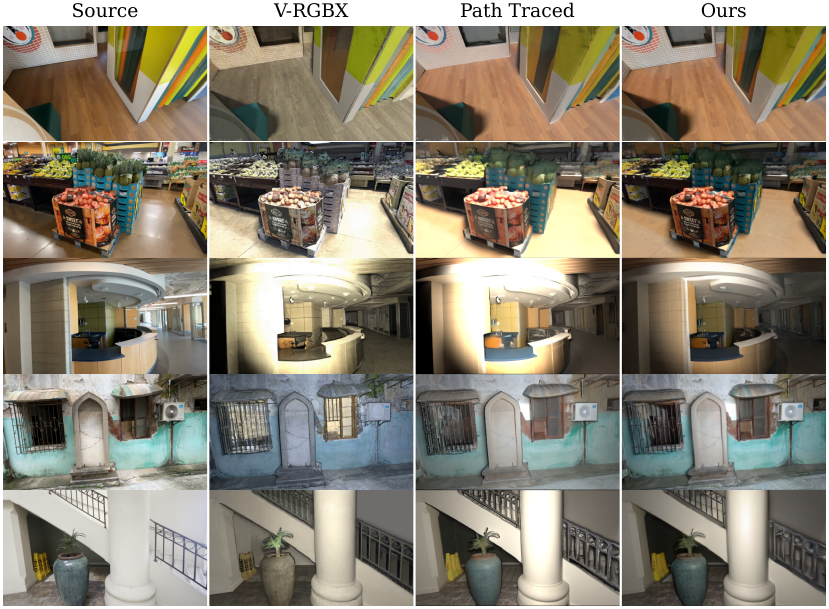}
    \caption{\textbf{Additional relighting comparisons on DL3DV scenes~\citep{dl3dv}.} Each row: a single source image, the most recent shading-conditioned baseline (V-RGBX), Blender's full RGB render of the reconstructed scene under the authored lighting (Path Traced), and \textsc{PIXLRelight}. \textsc{PIXLRelight} transfers the authored lighting while preserving the source's photographic detail.}
    \label{fig:dl3dv_qualitative_additional2}
\end{figure}

\clearpage
\section{Additional qualitative ablation results}
\label{sec:appendix:ablation}
\begin{figure}[!ht]
    \centering
    \includegraphics[width=\linewidth]{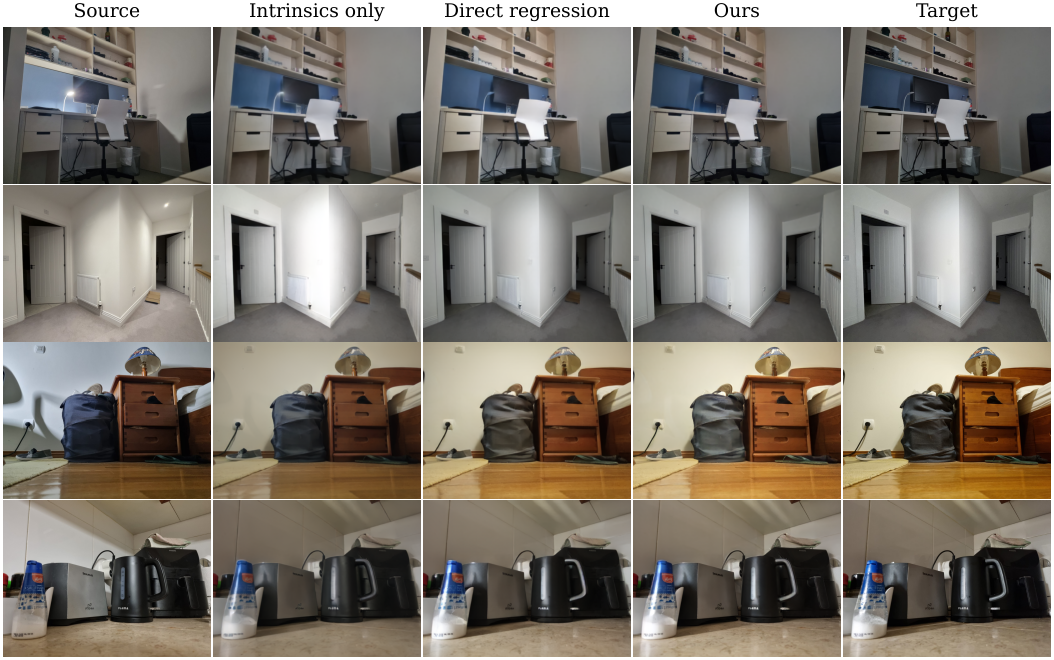}
    \caption{\textbf{Additional ablation comparisons on the held-out tripod captures.} Both variants are trained from scratch and differ from the full model in exactly one component. \emph{Intrinsics-only}: the source ViT branch is removed; the source enters the network only through the modulation head. \emph{Direct regression}: the modulation of~\cref{eq:modulation} is replaced by a sigmoid-activated RGB regression. Ours is the full model.}
    \label{fig:ablation_qualitative}
\end{figure}

\end{document}